\documentclass[11pt,a4paper]{article}
\usepackage[dvipsnames]{xcolor}
\usepackage[hyperref]{naaclhlt2019}
\usepackage{times}
\usepackage{latexsym}
\usepackage{todonotes}
\usepackage{url}
\usepackage{microtype}
\usepackage{xspace}
\usepackage{color}
\usepackage{verbatim}

\newcommand{\white}[1]{{\textcolor{white}{#1}}} 
\aclfinalcopy 
\def\aclpaperid{2119X-P6E8B6E4G7} 

\usepackage{booktabs}
\usepackage{graphicx}
\usepackage{enumitem}
\usepackage{tabularx}
\usepackage{subcaption}

\definecolor{dkgreen}{RGB}{0,130,0}
\definecolor{maroon}{rgb}{0.5, 0.0, 0.0}

\newcommand{\eat}[1]{}

\newcommand{\final}[1]{ #1}

\newcommand{\PM}{Post-Modifier\xspace}
\renewcommand{\pm}{post-modifier\xspace}
\newcommand{\pmg}{post-modifier generation\xspace}

\newcommand{\PMG}{Post-Modifier Generation\xspace}
\newcommand{\pms}{post-modifiers\xspace}

\newenvironment{itemizesquish}{
    \begin{list}
        {\labelitemi}
        {
            \setlength{\itemsep}{0em}
            \setlength{\topsep}{0em}
            \setlength{\parsep}{0em}
            \setlength{\labelwidth}{0.5em}
            \setlength{\leftmargin}{\labelwidth}
            \setlength{\itemindent}{0pt}
            \addtolength{\leftmargin}{\labelsep}  
        }
} {
    \end{list}
}

\newcommand{\name}{\texttt{PoMo}\xspace}

\newcommand{\blank}{\underline{\hspace{1cm}}}
{\list{}{\leftmargin=#1\rightmargin=#1}\item[]}%
{\endlist}

\title{PoMo: Generating Entity-Specific Post-Modifiers in Context}

\author{Jun~Seok~Kang$^1$, {\bf Robert~L.~Logan~IV$^2$},
{\bf Zewei~Chu$^3$}, {\bf Yang~Chen$^3$}, \\
{\bf Dheeru~Dua$^2$}, {\bf Kevin~Gimpel$^4$}, 
{\bf Sameer~Singh$^2$}, {\bf Niranjan~Balasubramanian$^1$}\\
  $^1$Stony Brook University, NY, USA\\
  $^2$University of California, Irvine, CA, USA \\
  $^3$University of Chicago, IL, USA\\
  $^4$Toyota Technological Institute at Chicago, IL, USA\\
  {\tt \{junkang,niranjan\}@cs.stonybrook.edu } \\
  {\tt \{rlogan,ddua,sameer\}@uci.edu } \\
  {\tt \{zeweichu,yangc1\}@uchicago.edu}\\
  {\tt \{kgimpel\}@ttic.edu } \\
  }

\begin{document}
\maketitle

\begin{abstract}

We introduce entity post-modifier generation as an instance of a collaborative writing task. 
Given a sentence about a target entity, the task is to automatically generate a post-modifier phrase that provides contextually relevant information about the entity. For example, for the sentence, ``Barack Obama, \underline{\hspace{1cm}}, supported the \#MeToo movement.'',  the phrase ``a father of two girls'' is a contextually relevant post-modifier.
To this end, we build \name, a post-modifier dataset created automatically from news articles reflecting a journalistic need for incorporating entity information that is relevant to a particular news event. \name\ consists of more than 231K sentences with post-modifiers and associated facts extracted from Wikidata for around 57K unique entities. We use crowdsourcing to show that modeling contextual relevance is necessary for accurate post-modifier generation.

We adapt a number of existing generation approaches as baselines for this dataset. Our results show there is large room for improvement in terms of both identifying relevant facts to include (knowing which claims are relevant gives a $>20\%$ improvement in BLEU score), and generating appropriate post-modifier text for the context (providing relevant claims is not sufficient for accurate generation). We conduct an error analysis that suggests promising directions for future research.

\end{abstract}

\section{Introduction}
\begin{figure}[t!]
    \centering
    \includegraphics[width=0.5\textwidth]{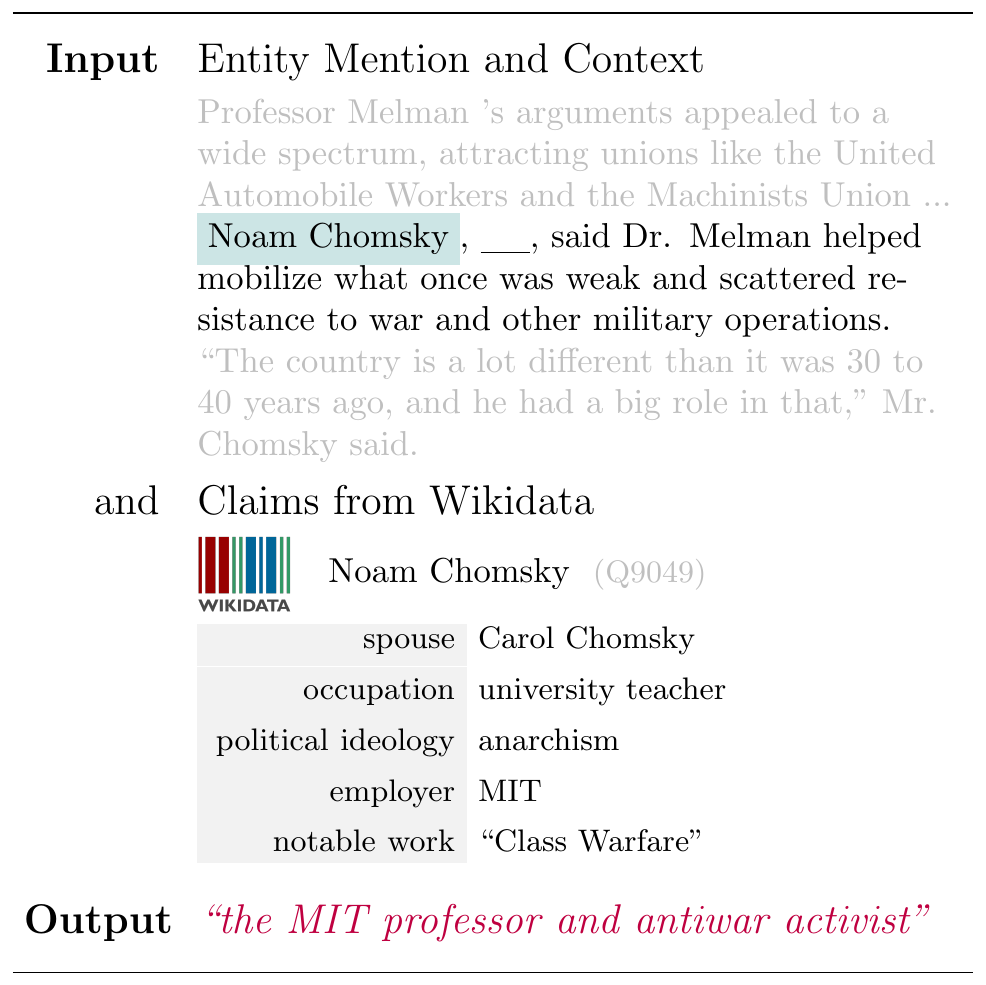}
    \caption{\PMG Task}
    \label{fig:post_modifier_generation}
\end{figure}

The goal of machine-in-the-loop writing systems is to assist human writers by directly augmenting their text. Examples include systems that refine human text for grammar~\cite{Rao2018DearSO}, collaborate on story plot generation systems~\cite{clark2018creative,Yu2012ASR}, or modify the content for style~\cite{hu2017toward,shen2017style,yang2018unsupervised}.
In this paper, we introduce \pmg as an instance of such an assistive writing task in the news domain. Journalists use \pms to introduce background information about entities discussed in news articles. To write these \pms journalists often need to look up relevant facts about entities. A \pmg system can be seen as a collaborative assistant that automatically finds relevant facts and inserts a small text fragment that augments the text produced by the human writer. 

Post-modifier generation is a \emph{contextual} data-to-text generation problem, where the data is the set of known facts about the target entity, and the text to be generated is a post-modifier that is relevant to the rest of the information conveyed in the text. Figure~\ref{fig:post_modifier_generation} shows an example. Given a sentence about the anti-war resistance work of Noam Chomsky, the target entity, and a set of known facts about him, the task is to generate a post-modifier that introduces Chomsky as a professor and mentions his background as an anti-war activist. An effective \pmg system must: (i) select suitable facts about the entity given the text, and (ii) produce text that covers these facts in a way that fits in with the rest of the text. 

We introduce \name, an automatically generated dataset for developing post-modifier generation systems.\footnote{\url{https://stonybrooknlp.github.io/PoMo/}} \name is a collection of sentences that contain entity post-modifiers, along with a collection of facts about the entities obtained from Wikidata~\cite{vrandevcic2014wikidata}. We use a small number of dependency patterns to automatically identify and extract post-modifiers of entities in sentences. We then link the extracted entities with the entries in Wikidata. The resulting dataset has 231,057 instances covering 57,966 unique entities. Our analysis show that the post-modifiers often combine multiple facts and are specific to the sentential context. 

We conduct two sets of experiments that highlight the challenges in \pmg.
\textbf{(i) Claim Selection:} 
Given an input sentence, the first step in generating a post-modifier is to figure out which facts to use. 
We formulate this as a distantly-supervised ranking problem, where we train neural models that learn to identify relevant claims for a given sentence. 
These claim ranking models perform well when predicting the relevance of coarse-grained facts (e.g. occupation), but fare poorly when predicting finer-grained facts (e.g. place of birth).
\textbf{ (ii) Generation:}
We adapt recent sequence-to-sequence generation models for this task. 
Results show that generation remains a challenge. 
Even though our automatic claim ranking does not improve generation, further experiments with oracle selected claims 
demonstrate that when relevant claims are known, the models can generate post-modifiers which humans deem comparable in quality to ones written by professional journalists.

In summary, the main contributions of this work are: 1) a data-to-text problem that introduces new challenges, 
2) an automated dataset creation pipeline and a large resulting dataset, 
3) a crowdsourcing study that  verifies the contextual relevance of post-modifiers, and 
4) a characterization of the difficulty of the task via performance analysis of numerous baselines.

\section{\name: Task and Dataset}

Post-modifier generation can be formulated as a data-to-text generation problem.
The input is text mentioning a target entity and a set of known facts about the entity.
The output is a phrase that: (i) fits as a post-modifier of the target entity mentioned in the input text, and (ii) conveys a subset of facts relevant to the context of the input text. 

Figure~\ref{fig:post_modifier_generation} shows an example for the target entity \texttt{Noam Chomsky}.
The input includes a sentence mentioning Chomsky's work on mobilizing anti-war groups along with its surrounding context, and a listing of all facts about Chomsky that are available in Wikidata.
Given these inputs, the task is to output a \pm phrase that conveys facts about Chomsky that fit within the sentence. 
In this example the \pm conveys both general background information about Chomsky (his occupation), and specific information relevant to the context of the sentence (being an anti-war activist).

This task can be seen as an instance of collaborative writing, where the journalist writes text about specific news events involving entities, and the generation system assists the journalist by inserting new text that augments the story.
Given a large collection of news articles, we can automatically create training data for such systems by removing the pieces of text that we want the assistant to generate.
This requires reliable ways to identify text to remove and sources of information that can be used to generate the text.
Here we describe a pipeline for generating such a dataset for our task. 

\subsection{Dataset}

We construct the \name dataset using three different news corpora: NYTimes ~\cite{sandhaus2008new}, CNN and DailyMail~\citep{hermann2015teaching}. 
We use Wikidata to collect facts about entities.\footnote{Wikidata dump from \url{https://www.wikidata.org/wiki/Wikidata:Database_download} (Dump date: 2018/06/25)}

\begin{table}[tb]
    \small
    \centering
    \begin{tabular}{rrrrr}
    \toprule
          & CNN   & DM    & NYT   & \textbf{Total} \\
    \midrule
    Train &        6,557  &      11,323  &   202,735  & \textbf{220,615 } \\
    Valid &            162  &            267  &        4,771  & \textbf{5,200 } \\
    Test  &            181  &            288  &        4,773  & \textbf{5,242 } \\
    \midrule
    Total  &        6,900  &      11,878  &   212,279  & \textbf{  231,057 } \\
    \bottomrule

    \end{tabular}%
  \caption{Dataset distribution by sources. }
  \label{tab:data_sources}
\end{table}%

\subsubsection{\PM and Entity Identification}
We use Stanford CoreNLP~\cite{manning-EtAl:2014:P14-5} to parse each sentence in the news articles and to identify named entities. 
We extract post-modifiers by finding noun phrases that share an \emph{appos} relation\footnote{An \emph{appos}itional modifier of an NP is another NP immediately to the right that defines or modifies the NP.}
with any recognized named entity in the sentence. 
In this work, we only consider post-modifiers for \emph{people}. 
In the future, we plan to expand \name to include more post-modifiers for other targets, such as organizations.
We extract only one such pair from a given sentence to reduce the possible noise in the extraction process. 

In our running example from Figure~\ref{fig:post_modifier_generation}, \texttt{Noam Chomsky} is recognized as a person entity.
The word ``professor'' is an appositive dependency of the word ``Chomsky'' and therefore, we extract the NP ``the Massachusetts Institute of Technology professor and antiwar activist'' which includes the word ``professor'' as a \pm for the target entity \texttt{Noam Chomsky}.

\subsubsection{Entity Claim Matching}
\label{subsec:entity_linking}

Wikidata provides information about entities in the form of key-value pairs that are called \emph{claims}.
To collect the facts about a target entity, we need to link the target to a specific entity in Wikidata.
We first search through Wikidata labels and aliases to find candidates with the same name as the target.
We sort the candidates based on the number of claims that have a significant word overlap with the extracted post-modifier.
We  link the entity to the highest ranked candidate whose claims cover at least 30\% of the non stop words in the post-modifier.
If such a candidate is found we 
record the claims that overlap with the post-modifier.
If no such candidate is found then we discard the entity.

We evaluate this simple heuristic by comparing the results to using an off-the-shelf entity linking system AIDA-light~\cite{nguyen2014aida} and show the results in Table \ref{tab:aida_result}. 
We find that AIDA-light agrees with our entity linking in 91.2\% of the cases. 
AIDA-light is able to link 94.3\% of the entities we found from NYTimes, but for CNN and DailyMail, it links only 87.0\% and 86.34\% of the entities, respectively. 
This decrease is likely due to the fact that AIDA-light was last updated in 2014 while the CNN/DailyMail datasets contain articles collected until the end of April 2015. 
On the other hand, NYTimes articles range from 1987 to 2007.
Our heuristic seems to be reasonably reliable as it does not depend on anything else but the data sources: news articles and Wikidata.

\begin{table}[tb]
  \centering
    \small
    \begin{tabular}{lcc}
    \toprule
          & \multicolumn{1}{c}{\bf AIDA Succ.} & \multicolumn{1}{c}{\bf Agreement}\bf \\
          \midrule
    Overall & 93.66 & 91.22 \\
    \addlinespace
    Train & 93.65 & 91.16 \\
    Valid & 94.06 & 91.13 \\
    Test  & 93.80 & 93.65 \\
    \addlinespace
    CNN   & 87.03 & 90.34 \\
    DM    & 86.34 & 85.66 \\
    NYT   & 94.29 & 91.53 \\
    \bottomrule
    \end{tabular}
  \caption{Percent agreement with AIDA-light's named entity disambiguation results.}
  \label{tab:aida_result}
\end{table}%

\subsection{Analysis}

Table \ref{tab:data_sources} shows the distribution of the data sources over train, validation, and test sets.
All splits maintain the relative distributions of the data sources to prevent stylistic mismatches from influencing generation. 
We also ensure that there is no entity overlap among the splits.
Within the NYTimes data, we verify that the distribution over years between 1987 and 2007 is also similar over the sets. 

\paragraph{Distribution of Post-Modifiers and Entities}
\label{subsubsec:dist_pm_entities}

\begin{figure*}[t!]
    \centering

    \begin{subfigure}[t]{0.32\textwidth}
        \centering
        \includegraphics[scale=0.31]{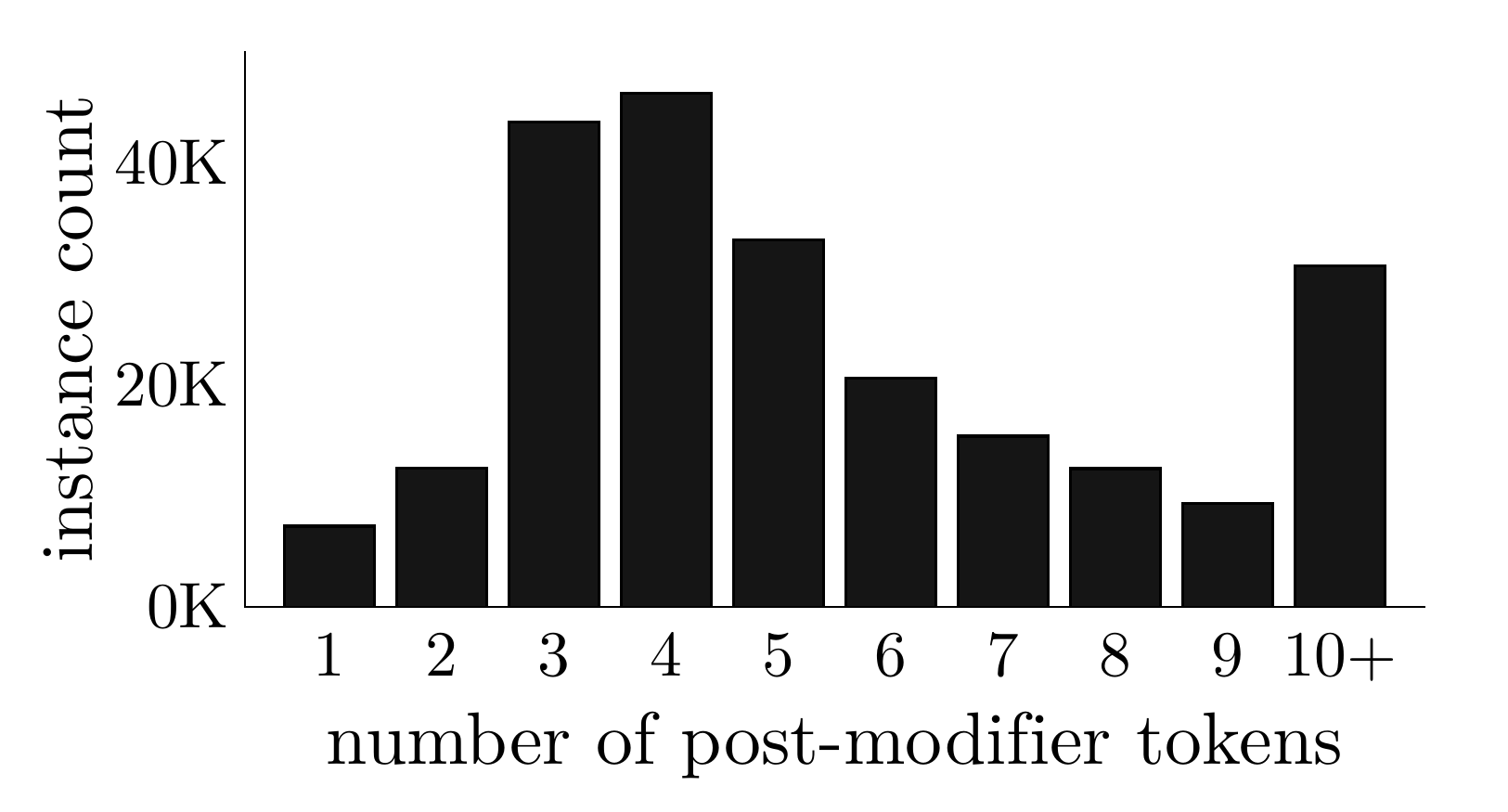}
        \caption{Histogram of the token counts of the \pms. Majority of the \pms (171K instances, 74.14\%) have 3 to 8 tokens. Average is 5.8 tokens.} 
        \label{fig:post_modifier_lengths}
    \end{subfigure}
    ~
    \begin{subfigure}[t]{0.32\textwidth}
        \centering
        \includegraphics[scale=0.31]{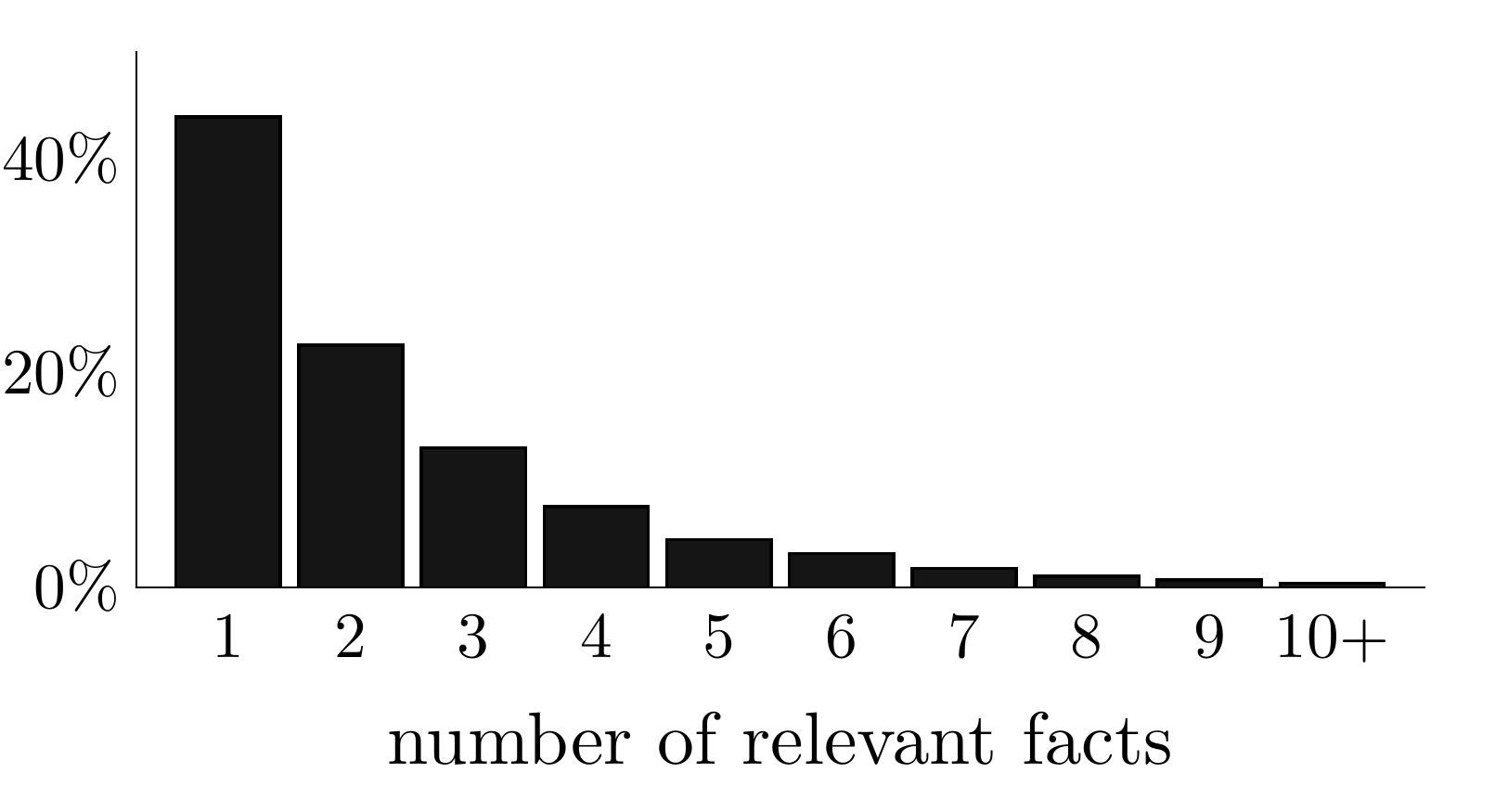}
        \caption{Number of relevant facts per instance in the dataset. More than a half of the \pms are related to two or more facts.
        }
        \label{fig:dataset_claim_coverage}
    \end{subfigure}
     ~
    \begin{subfigure}[t]{0.32\textwidth}
        \centering
        \includegraphics[scale=0.31]{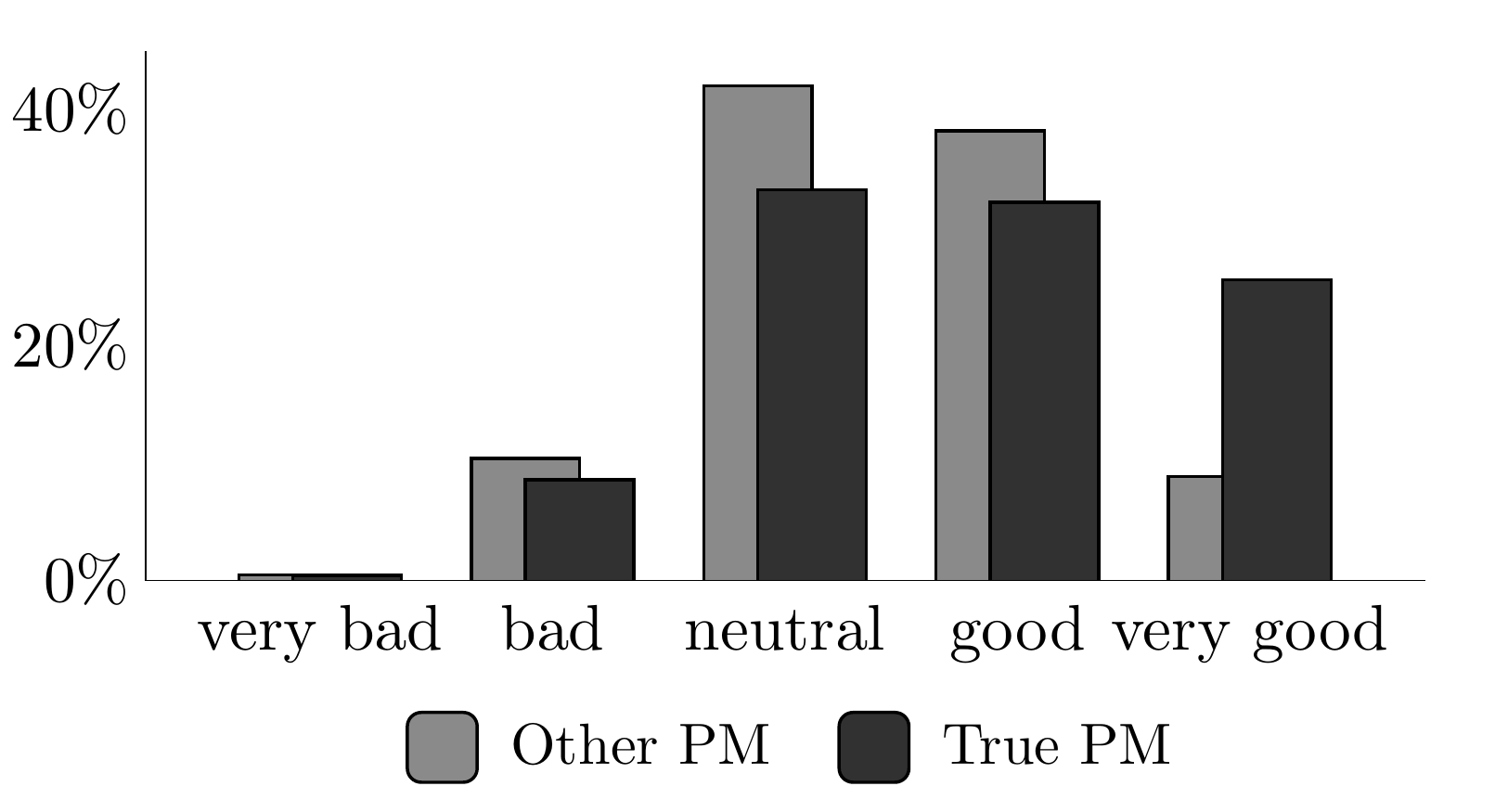}
        \caption{Histogram of the scores for \pms, averaged over three annotations.  The distribution of ratings for true and other post-modifiers. 
        }
        \label{fig:pm_score_histogram}
    \end{subfigure}
    \vspace{-0.2cm}
    \caption{\name Post-Modifier Statistics}
\end{figure*}

Figure~\ref{fig:post_modifier_lengths} 
shows the distribution of post-modifier lengths in terms of token counts.
Most post-modifiers are three to eight words long, and about $17.3\%$ are even longer.
Figure~\ref{fig:dataset_claim_coverage} shows an estimate of 
the number of relevant facts covered by the post-modifiers; this estimate uses the number of claims that overlap with the post-modifier via heuristic matching. 
More than half of the \pms convey two or more facts. About $11.4\%$ 
convey five or more facts.
These results suggest that generating post-modifiers requires composing together multiple relevant facts.

Table~\ref{tab:fact_type_hist} lists the most frequent types of facts used in the post-modifiers in our dataset.
Most relate to generic biographical information such as the entity's occupation, organizations they belong to, place of birth, etc.
Here again we see a range of types of information being conveyed which is likely to present a challenge for generation systems. 

The dataset also covers a wide variety of entity types.
We cluster the target entities by their occupation listed in Wikidata. We also use WordNet~\cite{miller1995wordnet} to traverse the hypernyms of the words to find frequent ones.
Then, we manually select the top ten occupation types.
Any entity that does not belong to the top ten is assigned to a single \emph{other} group.
The resulting distribution is shown in Table~\ref{tab:entity_type_distribution}.

\begin{table}[tb]
  \centering
    \small
    \begin{tabular}{r r}
    \toprule
    \textbf{Fact Type} & \textbf{Count} \\
    \toprule
    position held & 151,959 \\
    occupation & 82,781 \\
    educated at & 53,067 \\
    member of political party & 42,416 \\
    member of sports team & 41,602 \\
    employer & 36,412 \\
    award received & 31,618 \\
    position played on team / speciality & 23,987 \\
    country of citizenship & 17,444 \\
    nominated for & 15,139 \\
    place of birth & 9,185 \\
    participant of & 8,520 \\
    member of & 7,565 \\
    languages spoken,  written or signed & 4,827 \\
    place of death & 4,071 \\
    \bottomrule
    \end{tabular}%
  \caption{Top 15 frequent fact types based on heuristic fact coverage identification.}
  \label{tab:fact_type_hist}%
\end{table}%

\paragraph{Quality of Post-Modifiers} 

\begin{figure}[tb]
    \centering
    \includegraphics[width=\columnwidth]{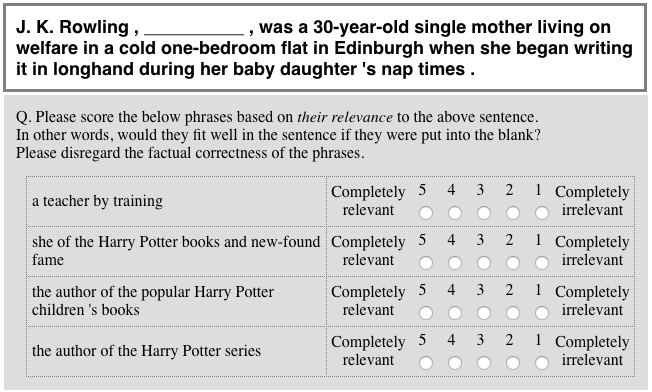}
    \caption{Screenshot of the crowdsourcing task. We asked crowd to rate the quality of post-modifiers.}
    \label{fig:amt_pm_score_screenshot}
\end{figure}

We conduct a crowdsourcing study to understand how often the \pms are specific to the particular context. 
For each (entity, context, post-modifier) triple in the validation set, we create multiple alternative post-modifiers by randomly choosing up to ten other post-modifiers that are found in some other sentences for the same entity. 
Crowd workers rate the quality of these post-modifiers. 
Figure~\ref{fig:amt_pm_score_screenshot} shows a screenshot of a task given to crowd workers. 
If the true \pm, the one that is actually used in the context, is rated the highest compared to the rest, then we assume the \pm is indeed specific to the context. 
On the other hand, if the crowd workers rate multiple other \pms as good fits for the context, then the true \pm is not context specific. 
Figure~\ref{fig:pm_score_histogram} shows the distribution of ratings for true and other \pms. The true \pms tend to be rated \emph{very good} or \emph{good} more often than the other \pms. This suggests that in many cases \pms are specific to the context and cannot be simply replaced by other \pms.

\begin{table}[tb]
  \centering
    \small
    \begin{tabular}{rrr}
    \toprule
    \textbf{Occupation} & \textbf{Count} & \textbf{Percentage} \\
    \toprule
    athlete & 13,560 & 23.39\% \\
    writer & 9,177  & 15.83\% \\
    politician & 8,518  & 14.69\% \\
    entertainer & 6,488  & 11.19\% \\
    \emph{other}  & 5,870  & 10.13\% \\
    scientist & 4,487  & 7.74\% \\
    artist & 4,175  & 7.20\% \\
    official & 2,098  & 3.62\% \\
    lawyer & 1,132  & 1.95\% \\
    educator & 961   & 1.66\% \\
    capitalist & 789   & 1.36\% \\
    scholar & 711   & 1.23\% \\
    \bottomrule
    \end{tabular}%
    \caption{Distribution of the inferred occupations of the target entities. Entities clustered by their occupation.
    }
  \label{tab:entity_type_distribution}%
\end{table}%

\section{Relevant Claim Selection}
\label{sec:claim_selection}

One of the key challenges of generating post-modifiers is to identify the claims about an entity that are relevant to the given context.
In this section, we explore methods for solving this task.

\subsection{Methods}

We consider three different models: a most-common claim baseline and two neural baselines.

\paragraph{Most-Common Claim}
This model employs a simple frequency heuristic: 
rank claims by the frequency of their types in the training post-modifiers (e.g. as in the order given in Table~\ref{tab:fact_type_hist}) and deem the top $n$ claims in this ranking as relevant. 

\paragraph{Neural Baselines}
We use two neural baselines with the following architecture.
Word embeddings are used to represent words in the context (e.g. current and previous sentence) and claims.
The sequences of embeddings are then fed through 2-layer LSTM's~\cite{hochreiter-lstm-1997} to obtain separate representations of the context and claims.
These representations are subsequently concatenated together and fed through a fully-connected layer with sigmoid activation, producing a scalar value for each claim representing the probability that it is relevant.
We use this model in two ways: as a classifier, and as a ranking model.
When used as a classifier, any claim whose score exceeds a threshold $\tau$ is predicted to be relevant.
When used as a ranking model, the top $n$ highest-scoring claims are predicted to be relevant.

\subsection{Experiments}

We train our baselines on the \name\ dataset, using the claims detected during dataset collection as a (distant) source of supervision.
Precision, recall, and $F_1$ score are used to evaluate model performance.
Model hyperparameters are chosen using (coarse) grid search to maximize $F_1$ score on the validation set.
The neural baselines use a vocabulary size of 50,000, 100-dimensional word embeddings, and 256 hidden units in the LSTM layers.
Dropout~\cite{dropout14} is applied between the LSTM layers with a $0.5$ keep probability. 
The neural classifier uses threshold $\tau=0.37$.
We find the optimal value of $n$ is $4$ for the most-common claims model and $2$ for the neural ranker.

Quantitative results are provided in Table~\ref{tab:claimrank_baselines}.
Both neural baselines perform considerably better than the most-common claims model.
This indicates that the provided contexts and claim values contain useful information for claim selection that goes beyond the information captured by global statistics of the dataset alone.
We additionally observe that the ranking-based approach outperforms the classification-based approach in terms of both precision and $F_1$ score, while having only slightly worse recall.

\begin{table}[!tp]
    \centering
    \small
        \begin{tabular}{lrrr}
        \toprule
                                        & \textbf{Prec.} & \textbf{Recall} & \boldmath$F_1$ \\
        \midrule
        Most-Common Claim ($n$=4)       & 39.9      & 51.6      & 45.0 \\
        Neural Classifier ($\tau$=0.37) & 52.0      & 63.8      & 57.4 \\
        Neural Ranker ($n$=2)           & 66.5      & 62.7      & 64.5 \\
        \bottomrule
        \end{tabular}%
    \caption{Baseline model performance on the claim selection task.}
    \label{tab:claimrank_baselines}
\end{table}%

To better understand the cases where the neural models fail and succeed, we examine the distribution of $F_1$ scores over the top 15 fact types (see Table~\ref{tab:f1_by_prop}).
Interestingly, when ranked by $F_1$ score we observe that fact types fall naturally into topically related groups:
\begin{enumerate}[nosep]
    \item position / occupation-related facts: \textit{position played, position held, occupation}
    \item membership-related facts: \textit{member of political party, member of, member of sports team}
    \item achievement-related facts: \textit{award received, nominated for}
    \item location-related facts: \textit{country of citizenship, place of death, place of birth}
\end{enumerate}
With the exception of \textit{employer}, the overarching trend is that the model identifies the relevance of coarse-grained claims better than fine-grained claims (e.g occupations, political parties, and sports positions are much more likely to be shared between entities than birth and death places).
This suggests that developing better methods for determining the relevance of fine-grained claims is a promising avenue for future research on this task.

\begin{table}[!tp]
    \centering
    \small
    \begin{tabular}{lr}
    \toprule
        \bf Fact Type & \boldmath$F_1$ \\
    \midrule
        employer & 76.95 \\
        position played on team / speciality & 76.65 \\
        position held & 63.10 \\
        occupation & 50.02 \\
        member of political party & 48.71 \\
        member of & 45.60 \\
        member of sports team & 38.53 \\
        award received & 37.53 \\
        nominated for & 30.87 \\
        educated at & 29.56 \\
        participant of & 29.04 \\
        country of citizenship & 16.28 \\
        place of death & 14.72 \\
        place of birth & 6.80 \\
        languages spoken, written or signed & 0.00 \\
    \bottomrule
    \end{tabular}
    \caption{$F_1$ score of neural ranker ($n=2$) on top 15 fact types.}
    \label{tab:f1_by_prop}
\end{table}

\section{Post-Modifier Generation}

We move our focus to the main task of post-modifier generation.

\subsection{Methods}
\label{sec:pm_methods}

At its core, post-modifier generation involves producing a variable-length sequence output conditioned on two variable-length inputs: the words in the current and previous sentence (e.g. the context), and the collection of claims about the entity.
Accordingly, the sequence-to-sequence (seq2seq) framework~\cite{sutskever2014} is a natural fit for the task --- we use it as the foundation for all of our baseline models.
Since research has shown that attention~\cite{bahdanau2015neural} and copy mechanisms~\cite{gu2016incorporating} consistently improve seq2seq model performance, we use these in our baselines as well.

\begin{figure}[t!]
    \centering
    \begin{subfigure}[t]{\linewidth}
        \centering
        \includegraphics[scale=0.70]{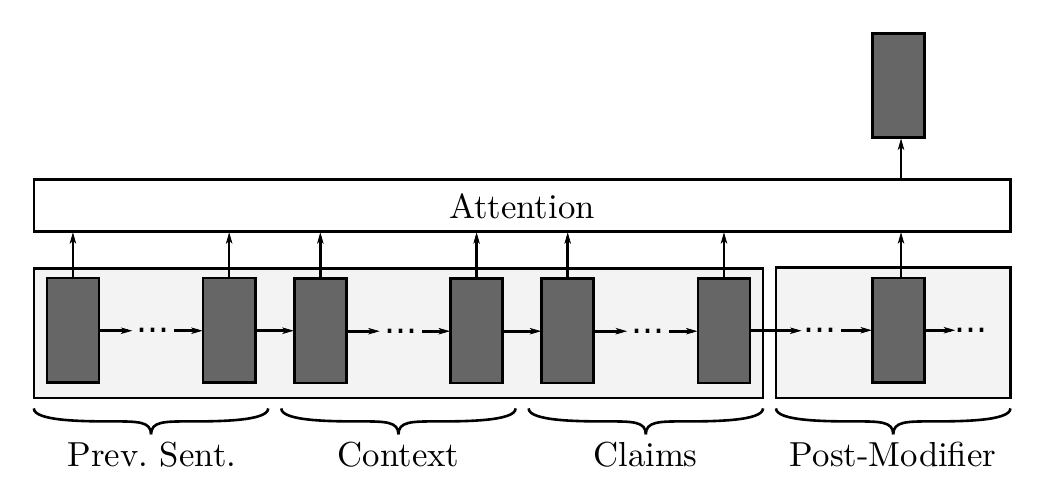}
        \caption{Basic sequence-to-sequence model} 
        \label{fig:seq2seq_1_encoder}
    \end{subfigure}

    \begin{subfigure}[t]{\linewidth}
        \centering
        \includegraphics[scale=0.70]{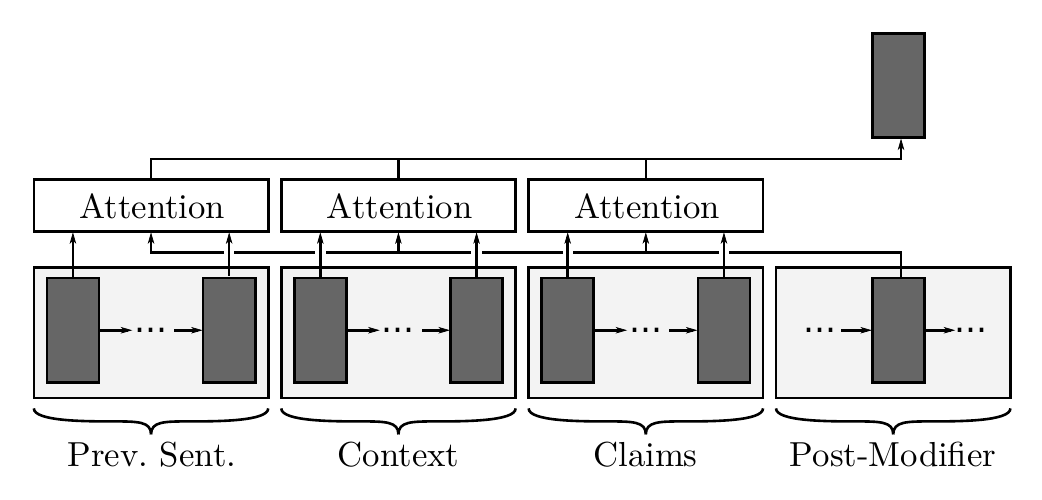}
        \caption{Tri-encoder model}
        \label{fig:seq2seq_3_encoders}
    \end{subfigure}

    \begin{subfigure}[t]{\linewidth}
        \centering
        \includegraphics[scale=0.70]{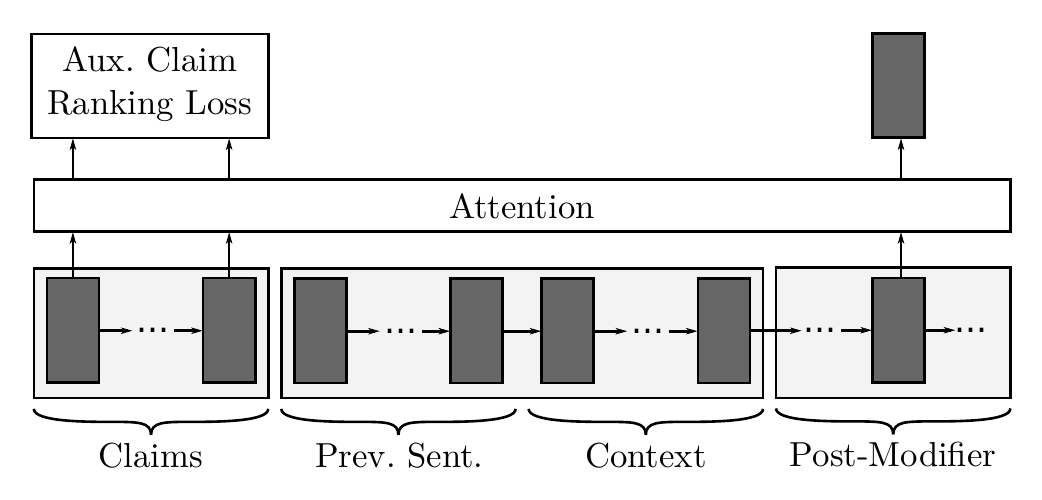}
        \caption{End-to-end claim selection model}
        \label{fig:e2e_claimrank}
    \end{subfigure}
    \caption{
        \name Models for post-modifier generation. 
        Grey boxes at the bottom represent individual encoder/decoder modules.
        (a) For baseline BiLSTM and transformer models all inputs are concatenated into one sequence.
        (b) The tri-encoder model has a separate encoder and attention for each type of input. The outputs of attention layers are concatenated together before generation. 
        (c) The end-to-end claim selection model attends to only the claim embeddings and uses an auxiliary loss term to encourage high attention scores for relevant claims.
    }
\end{figure}

One choice that must be made when using this framework is how to combine the different inputs.
The default approach we use is to concatenate the claim and context into a linear sequence of tokens during preprocessing (shown in Figure~ \ref{fig:seq2seq_1_encoder}).
We also experiment with encoding the claims and each of the context sentences separately, then concatenating their vector representations before decoding. We refer to this as the \textit{tri-encoder} approach (shown in Figure~\ref{fig:seq2seq_3_encoders}).  

As discussed earlier
, selecting relevant claims is crucial to generating good post-modifiers.
One way to incorporate claim selection is to use our baseline models from Section~\ref{sec:claim_selection} to cut out irrelevant claims from the input before feeding them to the encoder (e.g.\ performing hard claim selection).
This pipelined approach is not differentiable, and can suffer from cascading errors.
An alternative way is to use the model's attention mechanism as a form of soft claim selection that attends only to the relevant claims.
The drawback of this approach is that it does not make use of the available claim annotations, which are an important source of supervision.

Building on these observations, we propose an \textit{end-to-end claim selection} model which incorporates an additional term to the loss function that encourages the claim-level attention probabilities to be higher for the identified relevant claims as shown in Figure~\ref{fig:e2e_claimrank}.
The process for computing this loss term works as follows.
We begin by summing together attention scores for tokens within claims to obtain a claim-level score.
These scores are then fed through a sigmoid activation function to obtain a soft claim selection probability.
For each claim, we measure the binary cross entropy between the predicted selection probability and a binary variable indicating whether or not the claim was identified as relevant.
The final loss term is the average of these binary cross entropies.
Note that we do not use a copy mechanism in this model to avoid double-counting (since relevant claims were identified using word overlap).

\subsection{Experiments}

We experiment with two types of encoder/decoder modules: bidirectional LSTMs, and transformers~\cite{vaswani17transformer}.
We use a vocabulary of size 50K, truncate the maximum input sequence length to 500, and use a batch size of 32 in all experiments.
To help models distinguish between claims and context we demarcate claim fields with special \texttt{<claim>}, \texttt{<key>}, and \texttt{<value>} tokens. 
We train all the models for 150k steps, and evaluate on the validation dataset every 10k steps.
Evaluation is performed using the BLEU~\cite{papineni2002bleu} and METEOR~\cite{banerjee2005meteor} translation metrics, and Precision, Recall and $F_1$ score of the predicted bag-of-words (omitting stopwords).
The model with the highest $F_1$ score on the validation set is used during test time.

For the bidirectional LSTM, we use 2 hidden layers with 512 hidden units, 500-dimensional word embeddings, and apply dropout between layers with a keep probability of 0.7.
Models are trained using stochastic gradient descent with a learning rate of 1.0. 
For the transformer model, we use 4 attention heads, 4 layers of transformer blocks with 64 hidden units for the encoder and the decoder, a penultimate hidden layer with 256 units, and 64-dimensional word embeddings.
Transformer models are trained using Adam~\cite{adam} with an initial learning rate of 2.0, and a label smoothing~\cite{Szegedy16labelsmoothing} factor of 0.1 when calculating loss.

\begin{figure}[t!]
    \centering
    \begin{subfigure}[t]{1\columnwidth}
        \centering
        \includegraphics[scale=0.5]{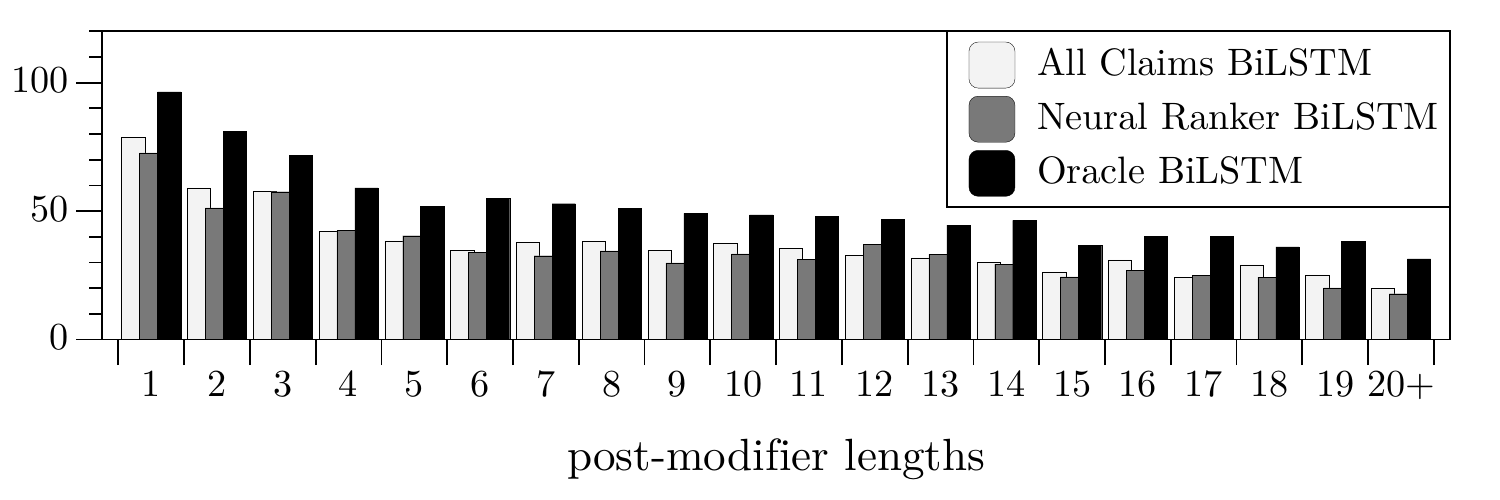}
        \vspace{-0.7cm}
        \caption{$F_1$ scores} 
        \label{fig:eval_pm_sizes_f1}
    \end{subfigure}
    \begin{subfigure}[t]{1\columnwidth}
        \centering
        \includegraphics[scale=0.5]{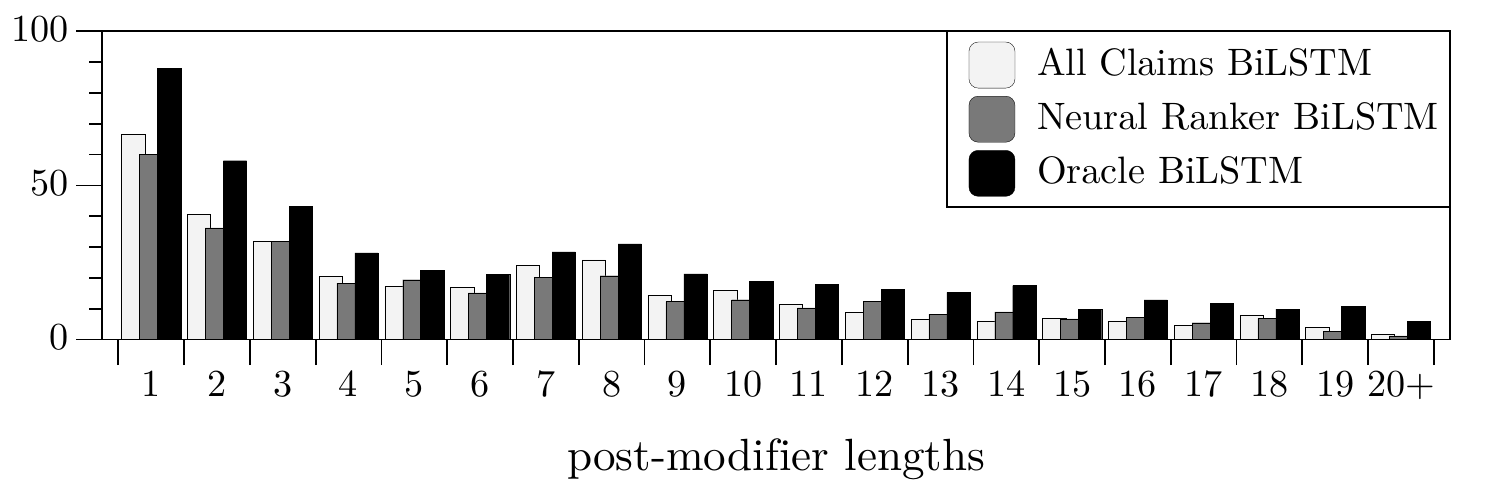}
        \vspace{-0.7cm}
        \caption{BLEU scores} 
        \label{fig:eval_pm_sizes_bleu}
    \end{subfigure}
    \caption{Performances by target post-modifier lengths of BiLSTM model. Post-modifiers with 20 or more tokens are put into one group, $20+$.}
    \label{fig:eval_pm_size}
    \vspace{-0.3cm}

\end{figure}

\begin{figure}[t!]
    \centering
    \begin{subfigure}[t]{1\columnwidth}
        \centering
        \includegraphics[scale=0.5]{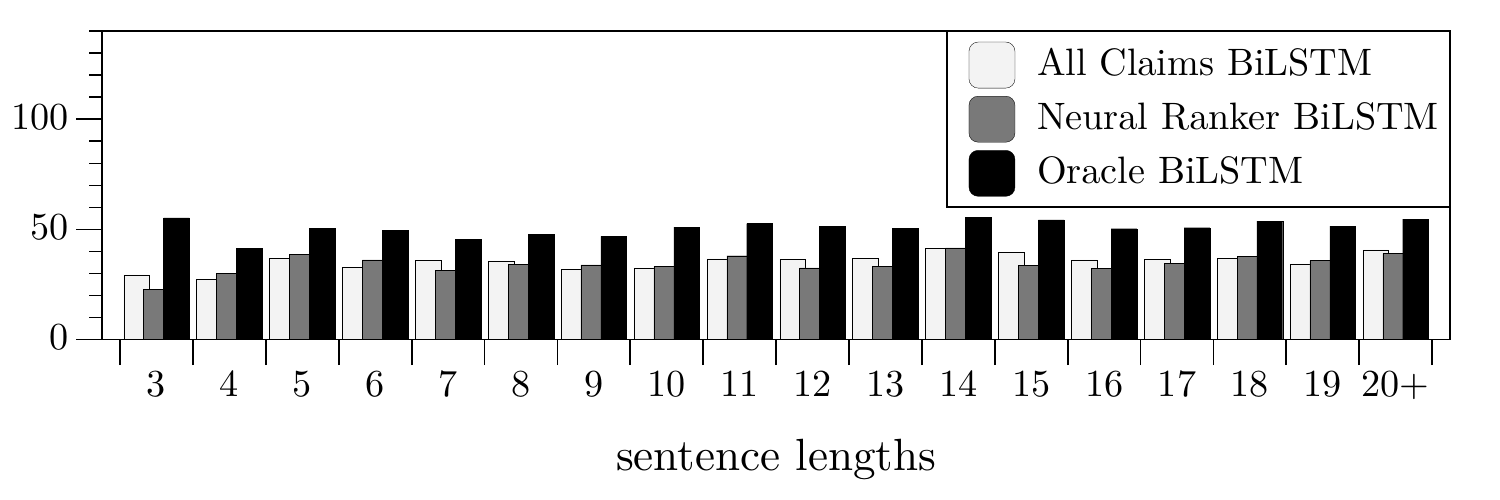}
        \vspace{-0.7cm}
        \caption{$F_1$ scores} 
        \label{fig:eval_sent_sizes_f1}
    \end{subfigure}
    \begin{subfigure}[t]{1\columnwidth}
        \centering
        \includegraphics[scale=0.5]{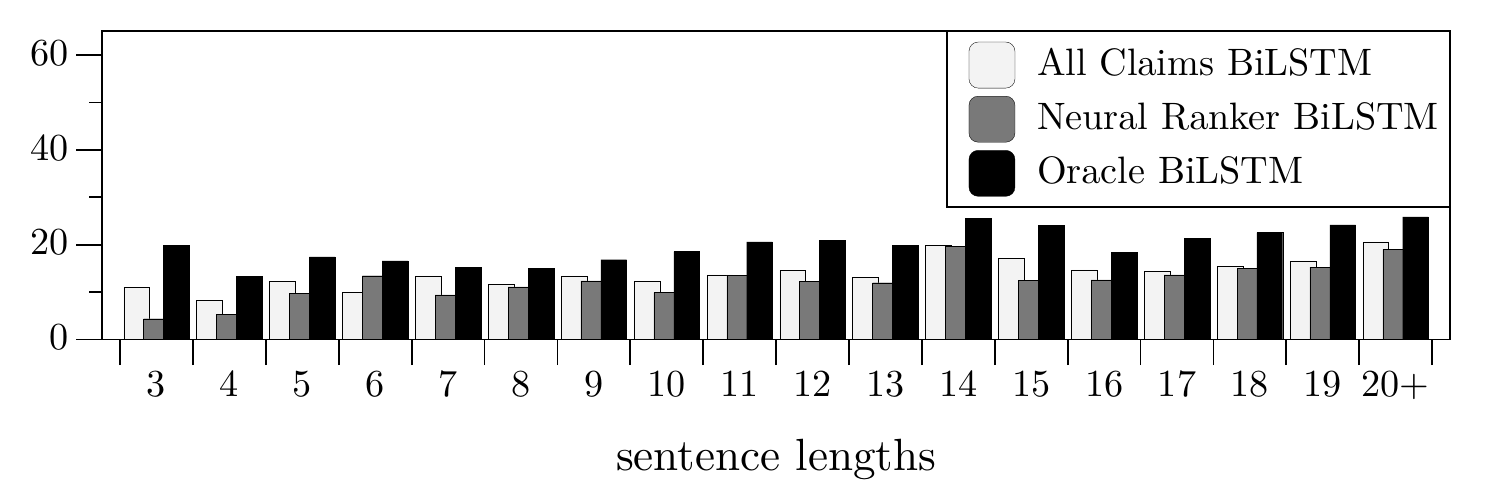}
        \vspace{-0.7cm}
        \caption{BLEU scores} 
        \label{fig:eval_sent_sizes_bleu}
    \end{subfigure}
    \caption{Performances by input sentence lengths of BiLSTM model. Sentences with 20 or more tokens are put into one group, $20+$.}
    \vspace{-0.3cm}
    \label{fig:eval_sent_sizes}
\end{figure}

We perform a variety of experiments, the results of which are displayed in Table~\ref{tab:pm_eval}.
In this table, \textit{Transformer} and \textit{BiLSTM} refer to models trained using the default approach to combining context and claims, while \textit{Tri-encoder} refers to a BiLSTM model trained using the approach described in~\ref{sec:pm_methods} (we do not train a transformer version since its performance is lackluster).
Here are detailed descriptions of the experiments performed in each section:
\begin{itemizesquish}
    \item \textbf{All Claims}: Results for vanilla seq2seq models.
    \item \textbf{Oracle}: Hard claim selection is performed using the oracle relevant claims.
    \item \textbf{Neural Ranker ($n=10$)}: Hard claim selection is performed using the top-10 claims returned by the neural ranker baseline.
    \item \textbf{End-to-End Claim Selection}: Results for the end-to-end claim selection model.
\end{itemizesquish}
In order to understand the relative contribution of the different inputs, we also include results for the BiLSTM model trained using either only the claims, or only the context sentences. \final{In Figure~\ref{fig:eval_pm_size} and \ref{fig:eval_sent_sizes}, we show the performances by \pm and sentence lengths to examine the impact of the such variables.}

\paragraph{Discussion of Quantitative Results}
Our results contain a few key findings.
The first is that knowing the relevant claims is critical to obtaining state-of-the-art performance; even knowing only oracle claims is sufficient to perform better than all of the other baselines, although there is a still a large improvement when context is additionally provided.
However, model-based approaches for claim selection do not seem to help: hard claim selection using the neural ranker performs just as well as the vanilla models, and our proposed approach for end-to-end claim selection has a negative impact.
This motivates the need for more effective methods of claim selection.
\final{The decreasing performances of the BiLSTM seq2seq models by the increasing target \pm and sentence lengths show the difficulty of generating long texts and handling long input data.}
Finally, we observe that the transformer-based seq2seq models are not particularly well-suited to this task. In all cases their performance is inferior to the BiLSTM-based approaches.
\final{Large-scale, pre-trained transformer-based language models, such as GPT-2~\cite{radford2019language} and BERT~\cite{devlin2018bert}, might be an interesting addition to the baselines, by framing the task as filling in the blanks for post-modifiers. However, when restricted to approaches that only use our dataset for training, we expect those based on language models to struggle due to the separation of entities among train, validation, and test.}

\begin{table}[t]
\small
\begin{center}
\begin{tabular}{lrrrrr}
\toprule 
 & \bf Prec. & \bf Rec. & \boldmath$F_1$ & \bf BLEU & \bf MET.  \\
 \midrule

\multicolumn{6}{l}{\bf All Claims} \\
Transformer & 41.9 & 22.2 & 29.0 & 7.0 & 12.1 \\ 
Tri-Encoder & 53.9 & 32.4 & 40.5 & 17.0 & 17.6 \\ 
BiLSTM  & 51.1 & 34.7 & 41.4 & 19.4 & 18.8  \\

\addlinespace
\multicolumn{6}{l}{\bf Oracle} \\
Transformer  & 69.4 & 38.6 & 49.6 & 15.7 & 20.0 \\ 
Tri-Encoder & 68.8 & 47.3 & 56.1 & 24.0 & 24.5\\ 
BiLSTM  &  66.4 & 48.8 & 56.2 & 25.1 & 25.3  \\ 

\addlinespace
\multicolumn{6}{l}{\bf Neural Ranker ($n=10$)} \\
Transformer & 41.5 & 22.4 & 29.1 & 6.9 & 12.1 \\ 
Tri-Encoder & 53.5 & 34.1 & 41.6 & 17.6 & 18.3 \\ 
BiLSTM  &  49.0 & 34.2 & 40.3 & 18.5 & 18.5 \\ 

\addlinespace
\multicolumn{6}{l}{\bf End-to-End Claim Selection} \\
BiLSTM  &  47.5 & 27.9 & 35.2 & 13.7 & 15.3 \\

\midrule
\multicolumn{6}{l}{\bf Context Only} \\
BiLSTM &  13.3 & 8.5 & 10.3 & 3.4 & 6.2 \\

\addlinespace
\multicolumn{6}{l}{\bf Claims Only} \\
BiLSTM & 47.3 & 28.5 & 35.6 & 13.5 & 15.0 \\

\addlinespace
\multicolumn{6}{l}{\bf Oracle Claims Only} \\
BiLSTM & 63.8 & 44.7 & 52.5 & 21.3 & 22.7 \\

\bottomrule
\end{tabular}
\end{center}
\vspace{-0.1cm}
\caption{
    Post modifier generation model performances with seq2seq models.
    Precision, recall and $F_1$ scores are computed ignoring stopwords.
}
\label{tab:pm_eval}
\vspace{-0.1cm}
\end{table}

\begin{table*}[!tb]
    \centering
    \small
    \begin{tabularx}{\linewidth}{r X}
    \toprule
        
        Input & Sky News reported Thursday night that Kenneth Clarke , \blank , had not yet decided whether to support Mr. Howard 's candidacy , raising the possibility the party could face a divisive battle for leadership . \\
        Claims 
            & + (position held: \textit{Chancellor of the Exchequer})\\
            & \white{+} (position held: \textit{Secretary of State for the Home Department}) \\
        Target & a former chancellor of the exchequer \\
        All Claims & the Home Secretary \\
        Oracle & the Chancellor of the Exchequer \\

    \midrule
        Input &`` A lot of people think it 's something we just started , but we actually opened the season with our first drive using it against Indianapolis , '' said Howard Ballard , \blank . \\
        Claims 
            & + (member of sports team: \textit{Buffalo Bills}) \\
            & + (position played on team / speciality: \textit{offensive tackle}) \\
            & \white{+} (mass: \textit{325 pound}) \\
            & \white{+} (height: \textit{78 inch}) \\
        Target & Buffalo 's robust , 6-foot-6-inch , 325-pound right tackle \\
        All Claims \& Oracle & the Bills ' offensive tackle \\

    \bottomrule
    \end{tabularx}
    \caption{
        \textbf{Challenging \name instances}.
        Two examples along with outputs of the best All Claims and Oracle models are displayed.
        Claims deemed relevant during dataset curation are prefaced with a +.
        In the first example, knowing the relevant claims helps the Oracle model produce an output that closely matches the Target, however lack of temporal information causes the model to miss the word \textit{former}.
        In the second example, the All Claims and Oracle models produce the same post-modifier.
        Although it is similar to the Target in meaning, it receives a low score using our evaluation metrics.
        Futhermore, our data curation method fails to identify relevant claims.
    }
    \vspace{-0.25cm} 
    \label{tab:pm_errors}
\end{table*}

\paragraph{Qualitative Analysis}

A cursory examination of model predictions (see Table~\ref{tab:pm_errors} for examples) provides insight into why post-modifier generation is a challenging task.
One issue that consistently appears is temporal inconsistency between the target and generated post-modifiers.
That is, the model may make an error since it is unaware of the time period that the article is written in (and also may not be aware of the periods of time for which a claim are true).
For example, in the first instance in Table~\ref{tab:pm_errors} the Oracle model predicts an almost correct post-modifier but misses the fact that Kenneth Clarke is a \textit{former} Chancellor of the Exchequer.
Another apparent issue is that models tend to generate shorter post-modifiers than humans.
As is indicated in Figure~\ref{fig:post_modifier_lengths} the post-modifiers in the dataset on average contain 5.8 tokens, whereas generated post-modifiers have only 3.8.
Lastly, we observe that our quantitative evaluation metrics can be too strict.
Take for example the second instance in Table~\ref{tab:pm_errors}.
Here the content of the target and generated post-modifiers is almost exactly the same, however our metrics would give very low scores due to low overlap.

\paragraph{Human Evaluation}

We additionally evaluate the generated post-modifiers by performing a human evaluation using Amazon Mechanical Turk. 
We randomly select 500 instances from test set and show crowdworkers the sentence context, along with the true \pm and a generated one.
For each instance, workers are asked to select the better phrase, or indicate that the two phrases are of equal quality.
For the Oracle BiLSTM model, the true \pms are preferred 46\% of the time, while generated \pms are preferred 43.2\% of the time. 
For the Neural Ranker ($n=10$) BiLSTM model, true \pms are favored much more (57.60\%) than the generated ones (20\%). Consistent with our quantitative results, we see that claim selection is a crucial factor in this task.
We also observe a few trends in the results.
People tend to prefer generated post-modifiers over the ones written by professional journalists when they are shorter and to use more general terms without elaborating too much about the entity.
In contrast, longer and more detailed human written \pms are preferred when they are especially relevant to the rest of the sentence.

\section{Related Work}

There is a large body of previous work on claim selection \cite{kukich1983design,duboue2003statistical,reiter1997building, tanaka1998reactive, barzilay2005collective} and language generation from structured data~\cite{reiter2005choosing, goldberg1994using}.
Initially, hand-crafted grammars were employed for language generation, which later evolved to statistical machine translation style models~\cite{wong2007generation} or PCFG based models~\citep{belz2008automatic}.
More recently, the focus has shifted to learning both fact selection and language generation jointly~\citep{liang2009learning,angeli2010simple,kim2010generative,lu2011probabilistic, konstas2013global}.

Modern approaches employ neural networks to solve this problem end-to-end.
\citet{mei2015talk} utilize an encoder-decoder framework to map weather conditions to a weather forecast. \citet{ahn2016neural} and \citet{yang2016reference} introduce a new class of language models which are capable of entity co-reference and copying facts from an external knowledge base.
Building upon these models, \citet{wiseman2017challenges} introduce an auxiliary reconstruction loss which use the hidden states of the decoder to recover the facts used to generate the text.
\citet{liu2018table} introduce a hierarchical attention model for fact selection, with the higher level focusing on which records in the table to select and the lower level focusing on which cells in a particular row to pay attention to.

In order to train complex neural models, the quest for larger datasets has become paramount.
\citet{lebret2016neural} introduce the WikiBio dataset containing Wikipedia articles of famous people and the corresponding infobox tables.
One drawback  of this dataset is that it is easily solved using template-based models.
To address this issue, \citet{wiseman2017challenges} introduce the ROTOWire dataset, which contains summaries of basketball games that are very long and syntactically diverse.
A comprehensive list of datasets is provided in Appendix~\ref{sec:append-data2text}. 

\section{Conclusions and Future Work}

Inspired by recent work on collaborative writing and data-to-text generation, we introduce post-modifier generation, a task that bridges the gap between these two fields.
The task is to generate a factual description of an entity which fits within the context of a human written sentence.
In order to promote research on this task we present \name, a large dataset of automatically extracted post-modifiers from news articles, aligned to the Wikidata knowledge graph.
We study the performance of numerous strong baseline models on this dataset, with a particular focus on the specific sub-task of claim selection.
Our results demonstrate that when relevant claims are known, sequence-to-sequence models are capable of generating post-modifiers which humans deem comparable in quality to ones written by professional journalists.
However, according to both quantitative metrics and human judgment, performance is much lower when models must determine for themselves which claims are relevant.
These experiments suggest plausible pathways to achieving human-level performance on this task that are both challenging and interesting problems for future research.

\section*{Acknowledgments}
We would like to thank the Toyota Technological Institute at Chicago for hosting the Workshop on Collaborative and Knowledge-Backed Language Generation which initiated the efforts for this project. 
The authors would also like to thank David Yarowsky, Jason Eisner, Kevin Duh, Kyle Gorman, and Philipp Koehn for feedback on early ideas for post-modifier generation.

\bibliographystyle{acl_natbib}
\bibliography{main}

\begin{thebibliography}{44}
\expandafter\ifx\csname natexlab\endcsname\relax\def\natexlab#1{#1}\fi

\bibitem[{Ahn et~al.(2016)Ahn, Choi, P{\"a}rnamaa, and Bengio}]{ahn2016neural}
Sungjin Ahn, Heeyoul Choi, Tanel P{\"a}rnamaa, and Yoshua Bengio. 2016.
\newblock A neural knowledge language model.
\newblock \emph{arXiv preprint arXiv:1608.00318}.

\bibitem[{Angeli et~al.(2010)Angeli, Liang, and Klein}]{angeli2010simple}
Gabor Angeli, Percy Liang, and Dan Klein. 2010.
\newblock A simple domain-independent probabilistic approach to generation.
\newblock In \emph{Proceedings of the 2010 Conference on Empirical Methods in
  Natural Language Processing}, pages 502--512. Association for Computational
  Linguistics.

\bibitem[{Bahdanau et~al.(2015)Bahdanau, Cho, and Bengio}]{bahdanau2015neural}
Dzmitry Bahdanau, Kyunghyun Cho, and Yoshua Bengio. 2015.
\newblock Neural machine translation by jointly learning to align and
  translate.
\newblock In \emph{3rd International Conference on Learning Representations,
  {ICLR} 2015, San Diego, CA, USA, May 7-9, 2015, Conference Track
  Proceedings}.

\bibitem[{Banerjee and Lavie(2005)}]{banerjee2005meteor}
Satanjeev Banerjee and Alon Lavie. 2005.
\newblock Meteor: An automatic metric for mt evaluation with improved
  correlation with human judgments.
\newblock In \emph{Proceedings of the acl workshop on intrinsic and extrinsic
  evaluation measures for machine translation and/or summarization}, pages
  65--72.

\bibitem[{Barzilay and Lapata(2005)}]{barzilay2005collective}
Regina Barzilay and Mirella Lapata. 2005.
\newblock Collective content selection for concept-to-text generation.
\newblock In \emph{Proceedings of the conference on Human Language Technology
  and Empirical Methods in Natural Language Processing}, pages 331--338.
  Association for Computational Linguistics.

\bibitem[{Belz(2008)}]{belz2008automatic}
Anja Belz. 2008.
\newblock Automatic generation of weather forecast texts using comprehensive
  probabilistic generation-space models.
\newblock \emph{Natural Language Engineering}, 14(4):431--455.

\bibitem[{Clark et~al.(2018)Clark, Ross, Tan, Ji, and
  Smith}]{clark2018creative}
Elizabeth Clark, Anne~Spencer Ross, Chenhao Tan, Yangfeng Ji, and Noah~A Smith.
  2018.
\newblock Creative writing with a machine in the loop: Case studies on slogans
  and stories.
\newblock In \emph{23rd International Conference on Intelligent User
  Interfaces}, pages 329--340. ACM.

\bibitem[{Devlin et~al.(2018)Devlin, Chang, Lee, and
  Toutanova}]{devlin2018bert}
Jacob Devlin, Ming-Wei Chang, Kenton Lee, and Kristina Toutanova. 2018.
\newblock Bert: Pre-training of deep bidirectional transformers for language
  understanding.
\newblock \emph{arXiv preprint arXiv:1810.04805}.

\bibitem[{Duboue and McKeown(2003)}]{duboue2003statistical}
Pablo~A Duboue and Kathleen~R McKeown. 2003.
\newblock Statistical acquisition of content selection rules for natural
  language generation.
\newblock In \emph{Proceedings of the 2003 conference on Empirical methods in
  natural language processing}, pages 121--128. Association for Computational
  Linguistics.

\bibitem[{Goldberg et~al.(1994)Goldberg, Driedger, and
  Kittredge}]{goldberg1994using}
Eli Goldberg, Norbert Driedger, and Richard~I Kittredge. 1994.
\newblock Using natural-language processing to produce weather forecasts.
\newblock \emph{IEEE Expert}, 9(2):45--53.

\bibitem[{Gu et~al.(2016)Gu, Lu, Li, and Li}]{gu2016incorporating}
Jiatao Gu, Zhengdong Lu, Hang Li, and Victor~OK Li. 2016.
\newblock Incorporating copying mechanism in sequence-to-sequence learning.
\newblock In \emph{Proceedings of the 54th Annual Meeting of the Association
  for Computational Linguistics (Volume 1: Long Papers)}, volume~1, pages
  1631--1640.

\bibitem[{Hermann et~al.(2015)Hermann, Kocisky, Grefenstette, Espeholt, Kay,
  Suleyman, and Blunsom}]{hermann2015teaching}
Karl~Moritz Hermann, Tomas Kocisky, Edward Grefenstette, Lasse Espeholt, Will
  Kay, Mustafa Suleyman, and Phil Blunsom. 2015.
\newblock Teaching machines to read and comprehend.
\newblock In \emph{Advances in Neural Information Processing Systems}, pages
  1693--1701.

\bibitem[{Hochreiter and Schmidhuber(1997)}]{hochreiter-lstm-1997}
Sepp Hochreiter and Jürgen Schmidhuber. 1997.
\newblock Long short-term memory.
\newblock In \emph{Neural Computation}.

\bibitem[{Hu et~al.(2017)Hu, Yang, Liang, Salakhutdinov, and
  Xing}]{hu2017toward}
Zhiting Hu, Zichao Yang, Xiaodan Liang, Ruslan Salakhutdinov, and Eric~P Xing.
  2017.
\newblock Toward controlled generation of text.
\newblock In \emph{Proceedings of the 34th International Conference on Machine
  Learning-Volume 70}, pages 1587--1596. JMLR. org.

\bibitem[{Kim and Mooney(2010)}]{kim2010generative}
Joohyun Kim and Raymond~J Mooney. 2010.
\newblock Generative alignment and semantic parsing for learning from ambiguous
  supervision.
\newblock In \emph{Proceedings of the 23rd International Conference on
  Computational Linguistics: Posters}, pages 543--551. Association for
  Computational Linguistics.

\bibitem[{Kingma and Ba(2015)}]{adam}
Diederik~P. Kingma and Jimmy Ba. 2015.
\newblock Adam: {A} method for stochastic optimization.
\newblock In \emph{3rd International Conference on Learning Representations,
  {ICLR} 2015, San Diego, CA, USA, May 7-9, 2015, Conference Track
  Proceedings}.

\bibitem[{Konstas and Lapata(2013)}]{konstas2013global}
Ioannis Konstas and Mirella Lapata. 2013.
\newblock A global model for concept-to-text generation.
\newblock \emph{Journal of Artificial Intelligence Research}, 48:305--346.

\bibitem[{Kukich(1983)}]{kukich1983design}
Karen Kukich. 1983.
\newblock Design of a knowledge-based report generator.
\newblock In \emph{Proceedings of the 21st annual meeting on Association for
  Computational Linguistics}, pages 145--150. Association for Computational
  Linguistics.

\bibitem[{Lebret et~al.(2016)Lebret, Grangier, and Auli}]{lebret2016neural}
R{\'e}mi Lebret, David Grangier, and Michael Auli. 2016.
\newblock Neural text generation from structured data with application to the
  biography domain.
\newblock In \emph{Proceedings of the 2016 Conference on Empirical Methods in
  Natural Language Processing}, pages 1203--1213.

\bibitem[{Liang et~al.(2009)Liang, Jordan, and Klein}]{liang2009learning}
Percy Liang, Michael~I Jordan, and Dan Klein. 2009.
\newblock Learning semantic correspondences with less supervision.
\newblock In \emph{Proceedings of the Joint Conference of the 47th Annual
  Meeting of the ACL and the 4th International Joint Conference on Natural
  Language Processing of the AFNLP: Volume 1-Volume 1}, pages 91--99.
  Association for Computational Linguistics.

\bibitem[{Liu et~al.(2018)Liu, Wang, Sha, Chang, and Sui}]{liu2018table}
Tianyu Liu, Kexiang Wang, Lei Sha, Baobao Chang, and Zhifang Sui. 2018.
\newblock Table-to-text generation by structure-aware seq2seq learning.
\newblock In \emph{Thirty-Second AAAI Conference on Artificial Intelligence}.

\bibitem[{Lu and Ng(2011)}]{lu2011probabilistic}
Wei Lu and Hwee~Tou Ng. 2011.
\newblock A probabilistic forest-to-string model for language generation from
  typed lambda calculus expressions.
\newblock In \emph{Proceedings of the Conference on Empirical Methods in
  Natural Language Processing}, pages 1611--1622. Association for Computational
  Linguistics.

\bibitem[{Manning et~al.(2014)Manning, Surdeanu, Bauer, Finkel, Bethard, and
  McClosky}]{manning-EtAl:2014:P14-5}
Christopher~D. Manning, Mihai Surdeanu, John Bauer, Jenny Finkel, Steven~J.
  Bethard, and David McClosky. 2014.
\newblock \href {http://www.aclweb.org/anthology/P/P14/P14-5010} {The
  {Stanford} {CoreNLP} natural language processing toolkit}.
\newblock In \emph{Association for Computational Linguistics (ACL) System
  Demonstrations}, pages 55--60.

\bibitem[{Mei et~al.(2016)Mei, Bansal, and Walter}]{mei2015talk}
Hongyuan Mei, Mohit Bansal, and Matthew~R. Walter. 2016.
\newblock \href {https://doi.org/10.18653/v1/N16-1086} {What to talk about and
  how? selective generation using lstms with coarse-to-fine alignment}.
\newblock In \emph{Proceedings of the 2016 Conference of the North American
  Chapter of the Association for Computational Linguistics: Human Language
  Technologies}, pages 720--730, San Diego, California. Association for
  Computational Linguistics.

\bibitem[{Miller(1995)}]{miller1995wordnet}
George~A Miller. 1995.
\newblock Wordnet: a lexical database for english.
\newblock \emph{Communications of the ACM}, 38(11):39--41.

\bibitem[{Nguyen et~al.(2014)Nguyen, Hoffart, Theobald, and
  Weikum}]{nguyen2014aida}
Dat~Ba Nguyen, Johannes Hoffart, Martin Theobald, and Gerhard Weikum. 2014.
\newblock Aida-light: High-throughput named-entity disambiguation.
\newblock \emph{LDOW}, 1184.

\bibitem[{Papineni et~al.(2002)Papineni, Roukos, Ward, and
  Zhu}]{papineni2002bleu}
Kishore Papineni, Salim Roukos, Todd Ward, and Wei-Jing Zhu. 2002.
\newblock Bleu: a method for automatic evaluation of machine translation.
\newblock In \emph{Proceedings of the 40th annual meeting on association for
  computational linguistics}, pages 311--318. Association for Computational
  Linguistics.

\bibitem[{Radford et~al.(2019)Radford, Wu, Child, Luan, Amodei, and
  Sutskever}]{radford2019language}
Alec Radford, Jeff Wu, Rewon Child, David Luan, Dario Amodei, and Ilya
  Sutskever. 2019.
\newblock Language models are unsupervised multitask learners.

\bibitem[{Rao and Tetreault(2018)}]{Rao2018DearSO}
Sudha Rao and Joel~R. Tetreault. 2018.
\newblock Dear sir or madam, may i introduce the gyafc dataset: Corpus,
  benchmarks and metrics for formality style transfer.
\newblock In \emph{NAACL-HLT}.

\bibitem[{Reiter and Dale(1997)}]{reiter1997building}
Ehud Reiter and Robert Dale. 1997.
\newblock Building applied natural language generation systems.
\newblock \emph{Natural Language Engineering}, 3(1):57--87.

\bibitem[{Reiter et~al.(2005)Reiter, Sripada, Hunter, Yu, and
  Davy}]{reiter2005choosing}
Ehud Reiter, Somayajulu Sripada, Jim Hunter, Jin Yu, and Ian Davy. 2005.
\newblock Choosing words in computer-generated weather forecasts.
\newblock \emph{Artificial Intelligence}, 167(1-2):137--169.

\bibitem[{Sandhaus(2008)}]{sandhaus2008new}
Evan Sandhaus. 2008.
\newblock The new york times annotated corpus.
\newblock \emph{Linguistic Data Consortium, Philadelphia}, 6(12):e26752.

\bibitem[{Shen et~al.(2017)Shen, Lei, Barzilay, and Jaakkola}]{shen2017style}
Tianxiao Shen, Tao Lei, Regina Barzilay, and Tommi Jaakkola. 2017.
\newblock Style transfer from non-parallel text by cross-alignment.
\newblock In \emph{Advances in Neural Information Processing Systems}, pages
  6830--6841.

\bibitem[{Srivastava et~al.(2014)Srivastava, Hinton, Krizhevsky, Sutskever, and
  Salakhutdinov}]{dropout14}
Nitish Srivastava, Geoffrey Hinton, Alex Krizhevsky, Ilya Sutskever, and Ruslan
  Salakhutdinov. 2014.
\newblock \href {http://jmlr.org/papers/v15/srivastava14a.html} {Dropout: A
  simple way to prevent neural networks from overfitting}.
\newblock \emph{Journal of Machine Learning Research}, 15:1929--1958.

\bibitem[{Sutskever et~al.(2014)Sutskever, Vinyals, and Le}]{sutskever2014}
Ilya Sutskever, Oriol Vinyals, and Quoc~V Le. 2014.
\newblock \href
  {http://papers.nips.cc/paper/5346-sequence-to-sequence-learning-with-neural-networks.pdf}
  {Sequence to sequence learning with neural networks}.
\newblock In Z.~Ghahramani, M.~Welling, C.~Cortes, N.~D. Lawrence, and K.~Q.
  Weinberger, editors, \emph{Advances in Neural Information Processing Systems
  27}, pages 3104--3112. Curran Associates, Inc.

\bibitem[{Szegedy et~al.(2016)Szegedy, Vanhoucke, Ioffe, Shlens, and
  Wojna}]{Szegedy16labelsmoothing}
Christian Szegedy, Vincent Vanhoucke, Sergey Ioffe, Jonathon Shlens, and
  Zbigniew Wojna. 2016.
\newblock \href {https://doi.org/10.1109/CVPR.2016.308} {Rethinking the
  inception architecture for computer vision}.
\newblock In \emph{2016 {IEEE} Conference on Computer Vision and Pattern
  Recognition, {CVPR} 2016, Las Vegas, NV, USA, June 27-30, 2016}, pages
  2818--2826.

\bibitem[{Tanaka-Ishii et~al.(1998)Tanaka-Ishii, Hasida, and
  Noda}]{tanaka1998reactive}
Kumiko Tanaka-Ishii, K{\^o}iti Hasida, and Itsuki Noda. 1998.
\newblock Reactive content selection in the generation of real-time soccer
  commentary.
\newblock In \emph{Proceedings of the 17th international conference on
  Computational linguistics-Volume 2}, pages 1282--1288. Association for
  Computational Linguistics.

\bibitem[{Vaswani et~al.(2017)Vaswani, Shazeer, Parmar, Uszkoreit, Jones,
  Gomez, Kaiser, and Polosukhin}]{vaswani17transformer}
Ashish Vaswani, Noam Shazeer, Niki Parmar, Jakob Uszkoreit, Llion Jones,
  Aidan~N Gomez, \L~ukasz Kaiser, and Illia Polosukhin. 2017.
\newblock \href
  {http://papers.nips.cc/paper/7181-attention-is-all-you-need.pdf} {Attention
  is all you need}.
\newblock In I.~Guyon, U.~V. Luxburg, S.~Bengio, H.~Wallach, R.~Fergus,
  S.~Vishwanathan, and R.~Garnett, editors, \emph{Advances in Neural
  Information Processing Systems 30}, pages 5998--6008. Curran Associates, Inc.

\bibitem[{Vrande{\v{c}}i{\'c} and Kr{\"o}tzsch(2014)}]{vrandevcic2014wikidata}
Denny Vrande{\v{c}}i{\'c} and Markus Kr{\"o}tzsch. 2014.
\newblock Wikidata: a free collaborative knowledgebase.
\newblock \emph{Communications of the ACM}, 57(10):78--85.

\bibitem[{Wiseman et~al.(2017)Wiseman, Shieber, and
  Rush}]{wiseman2017challenges}
Sam Wiseman, Stuart Shieber, and Alexander Rush. 2017.
\newblock \href {https://doi.org/10.18653/v1/D17-1239} {Challenges in
  data-to-document generation}.
\newblock In \emph{Proceedings of the 2017 Conference on Empirical Methods in
  Natural Language Processing}, pages 2253--2263, Copenhagen, Denmark.
  Association for Computational Linguistics.

\bibitem[{Wong and Mooney(2007)}]{wong2007generation}
Yuk~Wah Wong and Raymond Mooney. 2007.
\newblock Generation by inverting a semantic parser that uses statistical
  machine translation.
\newblock In \emph{Human Language Technologies 2007: The Conference of the
  North American Chapter of the Association for Computational Linguistics;
  Proceedings of the Main Conference}, pages 172--179.

\bibitem[{Yang et~al.(2017)Yang, Blunsom, Dyer, and Ling}]{yang2016reference}
Zichao Yang, Phil Blunsom, Chris Dyer, and Wang Ling. 2017.
\newblock \href {https://doi.org/10.18653/v1/D17-1197} {Reference-aware
  language models}.
\newblock In \emph{Proceedings of the 2017 Conference on Empirical Methods in
  Natural Language Processing}, pages 1850--1859, Copenhagen, Denmark.
  Association for Computational Linguistics.

\bibitem[{Yang et~al.(2018)Yang, Hu, Dyer, Xing, and
  Berg-Kirkpatrick}]{yang2018unsupervised}
Zichao Yang, Zhiting Hu, Chris Dyer, Eric~P Xing, and Taylor Berg-Kirkpatrick.
  2018.
\newblock \href
  {http://papers.nips.cc/paper/7959-unsupervised-text-style-transfer-using-language-models-as-discriminators.pdf}
  {Unsupervised text style transfer using language models as discriminators}.
\newblock In S.~Bengio, H.~Wallach, H.~Larochelle, K.~Grauman, N.~Cesa-Bianchi,
  and R.~Garnett, editors, \emph{Advances in Neural Information Processing
  Systems 31}, pages 7287--7298. Curran Associates, Inc.

\bibitem[{Yu and Riedl(2012)}]{Yu2012ASR}
Hong Yu and Mark~O. Riedl. 2012.
\newblock A sequential recommendation approach for interactive personalized
  story generation.
\newblock In \emph{AAMAS}.

\end{thebibliography}

\clearpage

\appendix

\section{Additional Claim Selection Materials}
\label{sec:append-claimrank}

Table \ref{tab:claimrank-mcc-val} lists the evaluation results of most-common claim baseline for $n$, the number of claims to predict, from 1 to 5. 
We obtain highest F1 with $n = 4$.

\begin{table}[htb]
    \centering
    \begin{tabular}{lrrr}
        \toprule
                $n$ & Precision & Recall    & F1 \\
        \midrule
                1   & 57.7      & 31.8      & 41.0 \\
                2   & 49.6      & 38.3      & 43.2 \\
                3   & 43.4      & 45.2      & 44.3 \\
                4   & 39.9      & 51.6      & 45.0 \\
                5   & 36.0      & 56.8      & 44.0 \\
        \bottomrule
    \end{tabular}
    \caption{Evaluation metrics for most-common claim baseline for different values of $n$.}
    \label{tab:claimrank-mcc-val}
\end{table}

Neural baseline shows improved performance compared to most-common claim baseline, showing its best performance when $n = 2$. 

\begin{table}[htb]
    \centering
    \begin{tabular}{lrrr}
        \toprule
                $n$ & Precision & Recall    & F1 \\
        \midrule
                1   & 75.2      & 42.4      & 54.3 \\
                2   & 66.5      & 62.7      & 64.5 \\
                3   & 56.0      & 69.6      & 62.1 \\
                4   & 48.7      & 76.2      & 59.4 \\
        \bottomrule
    \end{tabular}
    \caption{Evaluation metrics for neural baseline for different values of $n$.}
    \label{tab:claimrank-neural-val}
\end{table}

\begin{table}[tb]
    \centering
    \small
    \begin{tabularx}{\linewidth}{p{1.95cm}p{0.5cm}p{4.25cm}}
    \toprule
    \bf Dataset & \bf Size & \bf Domain of structured data to language\\
    \midrule
    WEATHER.GOV & 29.5k & Weather conditions to forecast report\\
    ALLRECIPES &  31k & Table of ingredients to recipes \\
    ROBOCUP & 1.5k & Game statistics to summaries\\
    ROTOWIRE & 4.9k & Basketball statistics to game summaries\\
    WIKIBIO & 728k &  Infobox to Wikipedia biography articles\\
    SBNations & 10.9K & Game statistic to fan written summaries \\
    WikiFacts & 40k & Freebase /film/actor facts to Wiki description of actor \\
    \bottomrule
    \end{tabularx}
    \caption{A comparative analysis of various datasets.    }
    \label{tab:dataset_comparison}
\end{table}

\section{Existing Data-to-Text Datasets}
\label{sec:append-data2text}

Table \ref{tab:dataset_comparison} provides a comprehensive list of  data-to-text datasets. 
\name presents a different set of challenges from these datasets. While the target text is shorter and less diverse, the task adds an additional challenge of figuring out which claims to use, a task which our evaluation shows is quite challenging.

\end{document}